%% file: universal.tex
\documentclass{article}

\usepackage{microtype}
\usepackage{graphicx}
\usepackage{subfigure}
\usepackage{booktabs} 

\usepackage{hyperref}

\usepackage{upquote} 
\usepackage{amssymb} 
\usepackage{mathtools} 

\usepackage{color}
\usepackage{listings}

\input{lstdodona}
\newcommand\ie{\emph{i.e.}\ }
\newcommand\eg{\emph{e.g.}\ }

\usepackage{array,etoolbox}
\preto\tabular{\setcounter{magicrownumbers}{0}}
\newcounter{magicrownumbers}

\usepackage[arxiv]{icml2020}

\icmltitlerunning{Universal Policies for Software-Defined MDPs}

\begin{document}


\twocolumn[
\icmltitle{Universal Policies for Software-Defined MDPs}



\icmlsetsymbol{equal}{*}

\begin{icmlauthorlist}
\icmlauthor{Daniel Selsam}{msr}
\icmlauthor{Jesse Michael Han}{upitt}
\icmlauthor{Leonardo de Moura}{msr}
\icmlauthor{Patrice Godefroid}{msr}
\end{icmlauthorlist}

\icmlaffiliation{msr}{Microsoft Research, Redmond, WA, USA}
\icmlaffiliation{upitt}{University of Pittsburgh, Pittsburgh, PA, USA}

\icmlcorrespondingauthor{Daniel Selsam}{daselsam@microsoft.com}

\icmlkeywords{search, heuristic, universal}

\vskip 0.3in
]



\printAffiliationsAndNotice{}  

\begin{abstract}
  We introduce a new programming paradigm called \emph{oracle-guided decision programming}
  in which a program specifies a Markov Decision Process (MDP) and the language provides
  a universal policy.
  We prototype a new programming language, \emph{Dodona}, that manifests this paradigm using
  a primitive \lstinline{choose} representing nondeterministic choice.
  The Dodona interpreter returns either a value or a \emph{choicepoint}
  that includes a lossless encoding of all information necessary in principle to make an optimal decision.
  Meta-interpreters query Dodona's (neural) \emph{oracle} on these choicepoints to get policy and value estimates,
  which they can use to perform heuristic search on the underlying MDP.
  We demonstrate Dodona's potential for zero-shot heuristic guidance by meta-learning over hundreds of synthetic tasks
  that simulate basic operations over lists, trees, Church datastructures, polynomials,
  first-order terms and higher-order terms.
\end{abstract}

\section{Introduction}\label{sec:intro}

One of the most encouraging discoveries in machine learning over the past several years
has been that simple neural network architectures can become meta-learners
when trained on sufficiently large datasets. This phenomenon is most apparent
in the language model \emph{GPT-3}~\cite{brown2020language}, which after having been
trained in an unsupervised fashion on a large corpora of text was able to perform
competently on many diverse language tasks.

While language models like GPT-3 can in principle be queried for guidance on any task that
boils down to emitting sequences of tokens, including software-definable tasks such as
program synthesis and theorem proving, language models suffer two stark and unnecessary limitations in software-definable domains.
The first problem is the \emph{specification problem}.
Although GPT-3 can perform competently on many diverse language tasks, querying the model
on a new task requires an \emph{ad-hoc} process whereby a human devises a \emph{prompt} such that the most likely
completion of the prompt on a random website may somehow include the desired information.
Some tasks do admit standard prompts---for example, ``TL;DR'' for requesting summarization---but
most tasks do not. For example, for a pre-trained language model to provide heuristic guidance deep inside a
theorem prover, somehow a lot of context must be provided to the network. What is to be proved? What is known or assumed?
What kind of prover is it?
What is the current state of the prover?
How does the current decision affect what the prover does downstream?
It may even be necessary to indicate explicitly in the prompt that one prefers a good decision, \ie one that will lead to finding a proof quickly.
Needless to say, all these descriptions might be intricate and awkward to elicit.
The second problem is the \emph{verification problem}. Since the tokens emitted by language models
cannot in general be trusted, they are only useful when it is easier to verify their outputs than to simply solve the problem from scratch.
This limitation is particularly salient in program synthesis as it is notoriously difficult to understand code one did not write.

In this work, our aim is to better leverage the tantalizing power of large-scale neural networks for problems that can be specified precisely.
Our main insight is that whereas the prompt is necessarily a dark art when natural language is involved,
we can construct a principled analogue of the idea for Markov Decision Processes (MDPs) that are defined by software.
For a given decision point within a software-defined MDP, the current state of the program along
with the code describing the downstream MDP together provide a precise representation of all that is necessary in principle
to make an optimal decision at the state in question. Thus a lossless encoding of this information can serve as a complete
and unambiguous prompt for a (neural) \emph{oracle}, that can provide a universal policy for all
computable MDPs. We show how such an encoding can be synthesized automatically, providing a solution
to the specification problem discussed above. The verification problem is trivially addressed by encoding the MDP precisely
in software in the first place.

Manifesting the framework just described requires a new kind of programming language whose programs define
MDPs rather than deterministic computations and whose runtime performs the bookkeeping
to synthesize the prompts at each decision point.
We refer to this broad paradigm as \emph{oracle-guided decision programming},
and prototype a new programming language, \emph{Dodona}, that manifests this paradigm.
Dodona---named after the oldest Hellenic oracle---is a minimalist, stand-alone language based on Scheme~\cite{sussman1998scheme}, designed to isolate the key ideas
of oracle-guided decision programming while avoiding unnecessary complexity.
\S\ref{sec:dodona} describes the basics of Dodona, both its semantics and how the interpreter
produces the datastructures that encode the relevant information at each choicepoint.
\S\ref{sec:embed} describes how we embed these datastructures into a graph-aware transformer~\cite{vaswani2017attention,hellendoorn2019global}.
This embedding represents only one small point in design space; there are many ways of embedding the information
but the trade-offs among them are not our present concern.
\S\ref{sec:experiments} describes an experiment where we generate data for hundreds of synthetic tasks
that simulate basic operations over lists, trees, Church datastructures, polynomials, first-order terms and higher-order terms;
we see that Dodona performs competently in the zero-shot setting on previously unseen tasks.
Finally, \S\ref{sec:discussion} includes extensive discussion of related work, missing features, and open problems.

\section{The Dodona Language}\label{sec:dodona}
\subsection{Quick tour}\label{subsec:dodona:tour}
The Dodona language is based upon a pure subset of the functional language Scheme~\cite{sussman1998scheme}, which is itself a minimalist dialect of Lisp~\cite{mccarthy1959lisp}.
Dodona is endowed with a \lstinline{choose} primitive that takes a list as argument and that evaluates \emph{nondeterministically} to one element of the list.
Thanks to \lstinline{choose}, a Dodona program defines a binary, deterministic Markov Decision Process (MDP), \ie a search problem,
where choosing from an empty list indicates failure.
The extension to intermediate, real-valued rewards and probabilistic transitions is relatively straightforward but not necessary for our present aims,
so we relegate discussion to \S\ref{sec:discussion}.
The syntax of Dodona is otherwise like Scheme:
\begin{lstlisting}
expr ::= c | x | (choose e)
| (lambda (x ...) e) | (if e1 e2 e3)
| (define x e) | (quote e)
| (e1 e2 ...)
\end{lstlisting}
where the prefix $c$ signifies a value (\eg a bool, int or list), $x$ a symbol and $e$ an expression.
We write \lstinline{(quote e)} as \lstinline{'e}, and it indicates that its argument should be returned
without being evaluated, so that \eg the expression \lstinline{'(0 1)} evaluates to the list containing 0 and 1.
The last form \lstinline{(e1 e2 ...)} represents function application. Dodona also includes standard
syntactic sugar, \eg \lstinline{(let ((x v)) b)} is sugar for \lstinline{((lambda (x) b) v)} and
\lstinline{let*} is sugar for nested \lstinline{let}s.

We illustrate Dodona with a few examples. We define \lstinline{fail} to be a function that chooses from an empty list:
\lstinline{(define (fail) (choose '()))}, and \lstinline{choose-bool} to be a function that chooses a bool:
\lstinline{(define (choose-bool) (choose '(#f #t)))}. Here is an expression that chooses a bool
and returns 7 if it is true and otherwise fails: \lstinline{(if (choose-bool) 7 (fail))}.
For our present purposes, all non-failing return values represent the same reward in the induced MDP.
Here is a higher-order, recursive program for producing lists:
\begin{lstlisting}
(define (choose-list choose-elem)
  (if (choose-bool) '()
      (cons (choose-elem)
            (choose-list choose-elem))))
\end{lstlisting}
Here is a higher-order, recursive program for producing trees:
\begin{lstlisting}
(define (choose-tree choose-leaf choose-node)
  (if (choose-bool)
      (choose-leaf)
      (let ((fn-nargs (choose-node)))
        (cons (first fn-nargs)
              (replicate (second fn-nargs)
                         (lambda ()
                           (choose-tree
                             choose-leaf
                             choose-node
                             )))))))
\end{lstlisting}
where \lstinline{choose-node} is expected to produce a two-element list (\lstinline{node-value}, \lstinline{num-args}),
and where \lstinline{(replicate n thunk)} returns a list of length \lstinline{n} with each element
generated by evaluating the thunk.

We remark that Dodona---like Scheme---may be minimalist but that it inherits the extreme power
and flexibility of the Lisp family.

\subsection{Evaluation} \label{subsec:dodona:eval}
Whereas a Scheme program evaluates to a value, a Dodona program evaluates to an (binary, deterministic) MDP.
We define our restricted notion of MDP as either a (terminal) value or a \emph{choicepoint},
where a choicepoint includes
(a) a list of possible values to choose from and
(b) a \emph{continuation} that takes whichever value may be chosen to another MDP.
The continuation in a choicepoint serves two roles in Dodona:
the \emph{computational} role, to enable continued evaluation for a particular choice,
and also the \emph{prompt} role, to represent all relevant information about the MDP in a way that
can be embedded. The latter role requires that the continuation be represented as a concrete datastructure,
as opposed to remaining implicit in the program state of the interpreter.

Dodona achieves this by representing a continuation as a stack of \emph{segments}, where a segment
is a triple consisting of: a syntax function, \ie a function in the host language that
maps a value to a Dodona expression, an environment to evaluate that expression in (a mapping from symbols to values), and the number of
arguments in the expression that will have already been evaluated.
Consider the following Dodona program:
\begin{lstlisting}
(let ((x 0))
  (+ x
     (let ((x 2)) (if (choose-bool) x 1))
     x))
\end{lstlisting}
The Dodona interpreter will evaluate this expression in applicative order until reaching the first (and only) choice,
and will return a choicepoint with choices \lstinline{(#f #t)} and continuation given by the following
stack of segments: [
(\lstinline{z} $\mapsto$ \lstinline{(if z x 1)}, \{\lstinline{x}:2\}, 1),
(\lstinline{z} $\mapsto$ \lstinline{(<prim:+> 0 z x)}, \{\lstinline{x}:0\}, 3)
].
In the second segment, the \lstinline{<prim:+>} represents the primitive procedure that \lstinline{+} evaluates to.
Note that in this segment, the first summand \lstinline{x} of the original expression has already evaluated to 0, whereas the last summand
\lstinline{x} has yet to be evaluated. This is reflected in the number 3 in the last field of the segment,
which indicates that three of the four arguments in the list \lstinline{(<prim:+> 0 z x)} have been evaluated already.

To summarize, the Dodona interpreter takes as input an expression, an environment, an integer indicating
the number of arguments that have already been evaluated, and a stack of segments, and returns as output
either a value or a choicepoint as described above.
We now present pseudocode for a few snippets of the interpreter.
The main function is \lstinline{step},
which dispatches to helper functions \lstinline{step-atom}, \lstinline{step-if},
\lstinline{step-quote}, \lstinline{step-lambda}, \lstinline{step-app} based on the
syntax of the expression it is evaluating.
Our actual implementation is tail-recursive, but we present a na\"{i}vely-recursive version here for simplicity.
To evaluate an atom, we first check if the continuation
stack is empty; if it is, we simply return the atom, and if it is not,
we pop the next segment from the continuation stack and continue evaluating:
\begin{lstlisting}
(define (step-atom x env i cstack)
  (if (null? cstack)
      x
      (let ((seg (first cstack)))
        (step ((seg->fn seg) x)
              (seg->env seg)
              (seg->idx seg)
              (rest cstack)
              ))))
\end{lstlisting}
To evaluate an if-statement, we first check if we have already evaluated the condition.
If not, we evaluate the condition while adding a segment to the continuation stack indicating
that we should resume evaluating the expression at argument index 1; if we have already
evaluated the condition,
we inspect it and evaluate either the then-branch or the else-branch accordingly:
\begin{lstlisting}
(define (step-if x env i cstack)
  (if (= i 0)
    (let ((f (lambda (y)
               (list 'if
                     y
                     (if->then x)
                     (if->else x)))))
      (step (if->cond x)
            env
            0
            (cons (seg->mk f env 1) cstack)))
    (let ((y (if (if->cond x)
                 (if->then x)
                 (if->else x))))
      (step y env 0 cstack))))
\end{lstlisting}

To evaluate a choicepoint, we first check if we have already evaluated the choices.
If not, we evaluate the choices while adding a segment to the continuation stack indicating
that we should resume evaluating the expression at argument index 1; if we have already
evaluated the choices, we simply return a choicepoint:
\begin{lstlisting}
(define (step-choicepoint x env i cstack)
  (if (= i 0)
      (let ((f (lambda (y)
                 (list 'choose y))))
        (step (second x)
              env
              0
              (cons (seg->mk f env 1) cstack)))
      (cp->mk (second x) cstack)))
\end{lstlisting}

The other cases in the interpreter follow the same pattern and so we omit them.

Our representation of a choicepoint as a list of choices and a stack of segments is clearly sufficient for its computational role.
On the other hand, the host-language syntax functions in the segments may seem problematic at first glance for the choicepoint's prompt role,
since we cannot easily inspect or embed a host-language function.
Our representation turns out to be very convenient for the prompt role as well:
when we build the graph of the choicepoint for the oracle, we apply these host-language functions to the
nodes that represent the values these functions will be applied to during evaluation.
We revisit this in \S\ref{sec:embed}.

\subsection{Meta-evaluation} \label{subsec:dodona:meta}
The Dodona interpreter will usually return a choicepoint as opposed to a value;
the power of Dodona comes from the meta-evaluators that we can write on top of the base interpreter. The
simplest is \lstinline{rollout}, which executes one path through the Dodona program, querying a callback \lstinline{decide} for
each decision that gets to observe the choicepoint:
\begin{lstlisting}
(define (rollout decide x env i cstack)
  (let ((y (step x env i cstack)))
    (if (value? y) y
        (let* ((idx (decide y))
               (choices (cp->choices y))
               (choice (nth idx choices))
               (cstack (cp->cstack y))
               (seg ((first cstack))))
          (rollout decide
                   ((seg->fn seg) choice)
                   (seg->env seg)
                   (seg->idx seg)
                   (rest cstack))))))
\end{lstlisting}
The \lstinline{rollout} procedure is illustrative but of little practical use since perfect guidance will be impossible for non-trivial domains.
In general it will be necessary to perform search, and it is also straightforward to write a best-first search procedure using a priority queue,
where the \lstinline{decide} callback is replaced by a \lstinline{score} callback that returns a probability distribution over the possible choices.
One can also implement more sophisticated heuristic search procedures such as Monte Carlo Tree Search (MCTS)
that make use of value-function estimates in addition to policy-function estimates.
Of course, heuristic search is only as good as the heuristic it employs.
The next section describes how we embed the choicepoints so that a neural oracle can be consulted for heuristic guidance.

\section{Embedding choicepoints}\label{sec:embed}

\subsection{Choicepoint graph}
We now explain how we embed the choicepoints returned by the evaluator (\S\ref{subsec:dodona:eval}).
The best way to embed software---be it code or intermediate representations---has been
under extremely active investigation for several
years and is far from settled~\cite{allamanis2017learning,wang2017dynamic,hellendoorn2017deep,ben2018neural,alon2018code2seq,alon2019code2vec,hellendoorn2019global,chen2019literature,wainakh2019evaluating,romanov2020representing,wang2020modular,chirkova2020empirical,brauckmann2020compiler,bieber2020learning,kanade2020learning,cummins2020programl,venkatakeerthy2019ir2vec,leather2020machine}.
Our goal in this work is to demonstrate the oracle-guided decision programming paradigm
of which the actual prompt-embedding is only one component, and a relatively modular one at that.
Thus we present a simple approach suitable for our preliminary experiments in Dodona,
and hope it is clear that more sophisticated approaches from the past or future
could be leveraged within the oracle-guided decision programming paradigm.

Recall from \S\ref{subsec:dodona:eval} that a choicepoint in Dodona consists of a list of values (\ie the choices)
and a stack of continuations, each one consisting of an environment (\lstinline{Env}) and a syntax function (\lstinline{Value -> Expr}).
Note that it is not trivial to embed this data since the environments may be enormous; it is as infeasible as it is pointless to embed
values in environments that are not used. Thus, some kind of analysis is required to find the subset of values in the environments that are
actually used. A compiled oracle-guided decision language may simply embed a lower-level representation of the prompt that already has its symbols resolved;
in Dodona, we simply perform this analysis online as we build the embedding.

We encode the choicepoint information by building a graph.
The encoding is relatively straightforward, so rather than belabor the low-level details we highlight a few decision decisions and then summarize the process.
\begin{itemize}
\item Every node stores a single token rather than arbitrary data, so that the initial embedding step can be performed
  efficiently on GPUs.
\item There are many different types of edges for different kinds of relationships between the nodes.
\item Every value is summarized by a single node no matter whether it is atomic and embeds to a single node (like a bool)
  or if it has internal structure and embeds to multiple nodes (like a list or a function). In particular, a function has a distinguished
  \textsc{function-summary} node that has (typed) incoming edges from its parameters and its body.
\item The procedure \lstinline{embed} takes an expression (where some terms may be nodes), an environment, and the number of arguments
  of the expression that have already been evaluated; it adds nodes and edges to the graph as appropriate, and returns the node summarizing its argument.
\item We unfold symbol-lookups recursively, but with memoization so that a single value in an environment is embedded at most once
  no matter how many times the symbol is referenced.
\item Lists are represented in terms of their \lstinline{cons} representation and so have linear depth,
  whereas unordered containers (\eg sets and maps) are embedded in a permutation-invariant way with constant depth.
\end{itemize}
To embed a choicepoint, we first embed all the choices and connect them to a distinguished \textsc{choice} node.
We then feed the \textsc{choice} node into the embedding
of the continuation stack. Specifically, for each segment in the stack, we apply the syntax function to the \emph{node}
representing the value that will be passed into it, and embed the result in the designated environment.
Note that a subset of the nodes correspond to the possible choices available at the choicepoint.

\subsection{Dynamic graph transformer (DGT)}\label{subsec:dodona:gnn}

The graph neural network (GNN) architecture that we find most intuitive does not seem to have been tried before,
so we introduce it here and call it the \emph{dynamic graph transformer} (DGT).
Like the \emph{GREAT} architecture~\cite{hellendoorn2019global}, the DGT uses the edge information to bias
the attention weights within a transformer encoder~\cite{vaswani2017attention}.
In GREAT, each attention layer has one scalar parameter for each edge type that indicates the bias for the edge type across all nodes and heads;
in particular, the edge bias is not conditioned on the nodes at all. In contrast, the DGT learns a
linear transformation for each attention layer that maps each node to a $\mathrm{numHeads} \times \mathrm{numEdgeTypes}$
matrix that represents the node's bias weight for each (attention head, incoming edge type) pair.
We implemented the DGT using PyTorch~\cite{paszke2019pytorch},
with the help of the PyTorch Lightning library~\cite{falcon2019pytorch}.

\section{Experiments}\label{sec:experiments}

No neural network of fixed size can evaluate arbitrary Dodona programs---not even
deterministic Dodona programs---since the language is Turing-complete
and hence evaluation may require arbitrarily large amounts of both memory and time.
Even Dodona programs that can be rolled out very efficiently
(as in \S\ref{subsec:dodona:meta}) may still require exponential time
to determine the optimal choice at a given choicepoint, \eg if the program
nondeterministically chooses a sequence of bits and then observes that those bits
invert a cryptographic hash of a known input sequence.
Of course, these limitations do not rule out the possibility of a useful oracle,
since many programs of interest require only moderate resources to evaluate, and
even exponentially-branching search spaces may permit extremely
effective heuristics.

We also note that by nature there can be no single ``killer application'' for a universal policy.
For any particular application with sufficiently large amounts of training data---and
with sufficiently motivated ML engineers---there will likely be advantages to designing
a custom representation and neural network that need not be isomorphic to the continuation
of the choicepoints. The promise of oracle-guided decision programming is in the potential gains in
statistical efficiency, \ie the amount of training data required for a new task.
The holy grail for our approach would be zero-shot learning for a wide range of important domains.
Based on the empirical findings from large-scale language models,
such meta-learning is likely to only be apparent when training at extremely large scales,
in terms of both the model complexity and the diversity of the
training data~\cite{DBLP:journals/corr/abs-2001-08361,DBLP:journals/corr/abs-2010-14701}.
While experiments at such large scales are not practically feasible for our first proof-of-concept,
we nevertheless demonstrate zero-shot learning over hundreds of synthetic tasks.

\paragraph{Tasks.}
We generate data for hundreds of synthetic tasks
that simulate basic operations over lists, trees, Church datastructures, polynomials, first-order terms and higher-order terms;
we see that Dodona performs competently in the zero-shot setting on previously unseen tasks.

Our workhorse is a simple program \lstinline{predict-app} in Dodona that generates synthetic data
for predicting the result of evaluating a deterministic Dodona program. Specifically, \lstinline{predict-app}
takes the following arguments:
\begin{itemize}
\item \lstinline{choose-func-arg}. A non-deterministic program returning a (\lstinline{fn}, \lstinline{arg}) pair.
\item \lstinline{max-results}. The number of (\lstinline{fn}, \lstinline{arg}) pairs to enumerate.
\item \lstinline{choose-output}. A non-deterministic program that can produce the results of
  evaluating the Dodona expression \lstinline{(fn arg)} for every (\lstinline{fn}, \lstinline{arg}) in the codomain of \lstinline{choose-func-arg}.
\item \lstinline{invert-output}. A program that inverts \lstinline{choose-output}, \ie that takes a value and produces
  the sequence of choices such that rolling out \lstinline{choose-output} using that sequence will produce the desired value.
\end{itemize}
The program \lstinline{predict-app} enumerates the first \lstinline{max-results} outputs of \lstinline{choose-func-arg};
for each (\lstinline{fn}, \lstinline{arg}) pair, it evaluates \lstinline{(fn arg)} to yield \lstinline{target},
and uses the result of \lstinline{(invert-output target)} as the correct sequence of choices for the Dodona program
\lstinline{(if (= (choose-output) (fn arg)) #t (fail))}.
Every call to \lstinline{predict-app} constitutes one task, and we set \lstinline{max-results} to be 500 for each task in the suite.
Note that this may induce more than 500 datapoints per task since each (\lstinline{fn}, \lstinline{arg})
pair may itself require many choices made in sequence, e.g. to produce a tree of digits.

Here is a sample of the tasks in our suite:
\begin{itemize}
\item \emph{Identity.} Predicting the result of applying the identity function to various values, including various
  types of lists and trees. These tasks are not quite as trivial as they may seem, since the network must be able to make
  the connection between the embedding of the values downstream, its partially constructed guess, and its current choices.

\item \emph{Arithmetic.} Predicting the result of evaluating different kind of (integer) arithmetic expression trees involving
  adding, subtracting, multiplying, maximums, minimums, dividing, and computing remainders.
  The tasks are parameterized by tree-generators, \eg one task might have the form \lstinline{(+ _ _)}
  while another may have the form \lstinline{(+ _ (* _ _))} where the underscores indicate nondeterministic
  integers.

\item \emph{Lists.} Predicting the result of many basic functions on lists of digits,
  such as computing the length,
  taking and dropping a specified number of elements,
  finding the $n$th element,
  determining if there exists an element that satisfies a certain property,
  counting, erasing and filtering the elements that satisfy certain properties,
  mapping simple unary functions, concatenating two lists, and others.
  The list tasks also include folding binary Boolean operations over lists of Boolean's.

\item \emph{Trees.} Predicting the result of many basic functions on trees (with all values at the leaves),
  such as computing the number of inner nodes,
  the number of leafs,
  the total number of nodes,
  the depth,
  the subtree at a specified path in the tree
  and the number of leaves satisfying certain properties,
  mapping simple unary functions over the tree,
  and folding simple binary Boolean operations over the leafs of Boolean trees.

\item \emph{Polynomials.} Predict the result of applying various components of the
  sparse Horner normal form computation~\cite{gregoire2005proving}
  to different classes of polynomials, \eg with different numbers of differently named variables.

\item \emph{First-order simplification.} Predict the result of simplifying (\ie exhaustively rewriting) first-order terms
  for a range of different simplification rules, \eg one task may involve simplifying \lstinline{(f ?x)} to \lstinline{(g ?x)}
  whereas another may involve simplifying \lstinline{(f (f ?x))} to \lstinline{(g (g ?x))} and
  \lstinline{(g (g ?x)} to \lstinline{(h ?x)}.

\item \emph{Higher-order reduction.} Predict the result of performing $\beta$ and $\eta$ reduction on higher-order terms
  with different sets of differently typed metavariables, as well as a few other higher-order tasks such as type inference,
  and term substitution followed by $\beta\eta$ reduction~\cite{dowek2001higher}.

\item \emph{Planted path.} Chart a path through some encoding of a binary tree to find the single non-failing leaf, where
  the tasks range over different encodings of trees and paths. For example, one task encodes binary trees using nested
  Church pairs, where a path indicates a sequence of Church pair projections.
\end{itemize}

The supplementary material includes code for all tasks in the suite.

\begin{figure*}
  \begin{center}
    \includegraphics[width=0.9\textwidth]{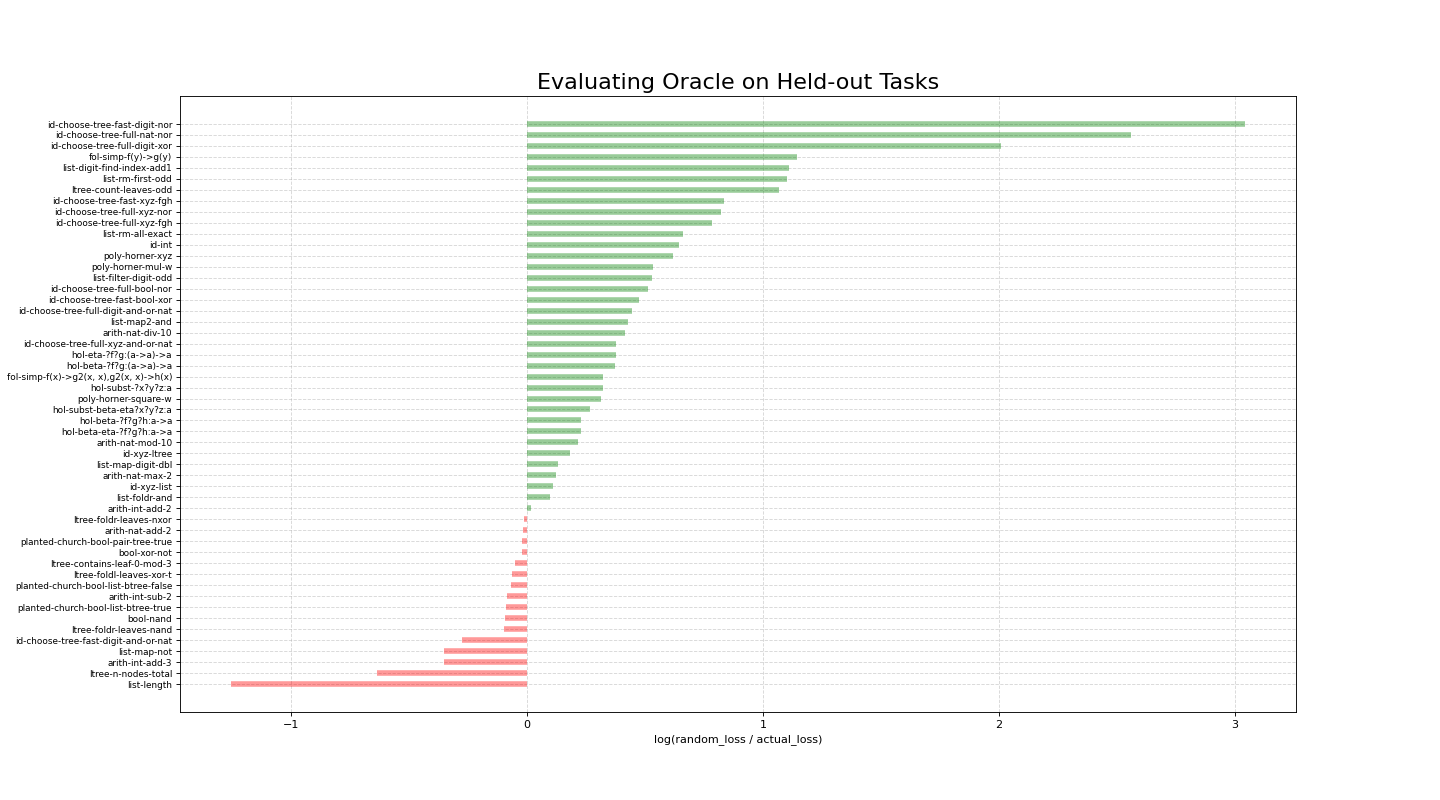}
    \caption{The results of the experiment described in \S\ref{sec:experiments}.
Each row corresponds to a held-out task. The horizontal bar for a task indicates the metric
$\log (\mathrm{uniformLoss}/\mathrm{actualLoss})$ where $\mathrm{uniformLoss}$ is the loss achieved by the uniform distribution over choices.
This metric is 0 when the trained network is as good as uniform guessing,
positive (and green) when it is better than uniform guessing, and negative (and red) when it is worse than uniform guessing.
The majority of tasks are in the green, often significantly so, indicating some degree of zero-shot learning.
}
    \label{fig:test-plot}
  \end{center}
\end{figure*}

\paragraph{Results.}
We randomly divided the tasks into a 70-10-20 train-valid-test split, and trained a 12-layer DGT (see~\S\ref{subsec:dodona:gnn})
on the training set using the validation set for early stopping.\footnote{Other hyperparameters: 128-dimensional embeddings, 4 attention heads each of dimension 32, and learning rate 1e-4.} The results on the held-out test set are shown in Figure~\ref{fig:test-plot}.
Each row corresponds to a held-out task. The horizontal bar for a task indicates the metric
$\log (\mathrm{uniformLoss}/\mathrm{actualLoss})$ where $\mathrm{uniformLoss}$ is the loss achieved by the uniform distribution over choices.
This metric is 0 when the trained network is as good as uniform guessing,
positive (and green) when it is better than uniform guessing, and negative (and red) when it is worse than uniform guessing.
The majority of tasks are in the green, often significantly so, indicating some degree of zero-shot learning.

However, we stress that the primary motivation for this experiment was to demonstrate the Dodona features discussed earlier in the paper
rather than to test a rigorous hypothesis. There is no rigorous motivation for the choice of tasks, and although we tried
to design the experiment so that we could not predict the outcome, countless design decisions in the
task suite could have made the resulting numbers higher or lower in myriad inscrutable ways.
Many tree-to-tree tasks such as first-order simplification and higher-order reduction may
leave most subtrees unchanged most of the time, potentially allowing the hypothesis that approximates the simplification
function with the identity function to beat random significantly. Even worse, many of the higher-order reduction tasks
are parameterized by the names and types of metavariables, but some of the terms generated may not include any variables at all;
this could cause some datapoints to be included in multiple nominally-different tasks.
On the other hand, the unseen task that the oracle performs worst on is \emph{list-length},
and this red would likely be green if \emph{list-length} were split into two tasks depending
on the type of elements in the list, \eg into a
\emph{list-digit-length} and \emph{list-bool-length}.
With these considerations in mind, we still consider the
predominance of green to be encouraging at the very least.


\section{Discussion}\label{sec:discussion}

Oracle-guided decision programming extends nondeterministic programming, which goes back at least to~\citet{mccarthy1959basis}
in which John McCarthy proposed the \lstinline{amb} (for ``ambiguous'') operator,
which is effectively the same as the \lstinline{choose} operator we use in Dodona.
\citet{zabih1987non} extended Scheme with \lstinline{amb} to produce the nondeterministic \emph{Schemer} language,
with a built-in notion of dependency-directed backtracking. This feature could be productively adopted by Dodona
and its descendants as well. \citet{siskind1993screamer} presented a nondeterministic
extension of Common Lisp called \emph{Screamer} while~\citet{andre2002state}
made explicit the connection between nondeterminism and Markov decision problems.
However, nondeterministic programming is of limited use in the absence of heuristics
since most nondeterministic programs of interest will be \emph{a priori} intractable
even with dependency-directed backtracking.
Our present work can be viewed in part as revisiting this classic work,
where we use machine learning to provide heuristic guidance.

Our work has many parallels with probabilistic programming languages as well
and is particularly inspired by the Scheme-based language, \emph{Church}~\cite{goodman2012church}.
Nondeterministic programming can in principle be simulated in probabilistic programming languages
by sampling all choices from a uniform distribution, implementing \lstinline{fail} as observing
an event of probability zero, and querying for the \emph{maximum a posteriori} (MAP) estimate over all the choices.
However, the inference algorithms adopted by probabilistic programming languages are not
designed for such ``hard'' observations and most---specifically the ones based on
local transitions---do not work at all on problems unless a random starting point is likely
to be connected to a solution by a path of nonzero probability.
Within probabilistic programming,
many projects have used neural networks to amortize the cost of inference within certain
families of models~\cite{ritchie2016neurally,le2017inference}, but none that we know of
have attempted domain-agnostic variants.
An analogue of our universal oracle could be devised for probabilistic programming languages
where instead of learning which decisions are preferable the oracle learns which distributions to propose
local transitions from.

The idea of improving sample efficiency by training on many tasks in parallel using a shared representation
was articulated in the early 1990's. \citet{baxter1995learning} suggested it as a new approach to machine learning:
``Instead of only learning the task required, learn as many related tasks as possible.''
\citet{caruana1995learning,caruana1997multitask} showed the efficacy of the regime empirically and coined the phrase \emph{multitask learning},
though Caruana credits~\citet{hinton1986learning} with the key insight that generalization may be improved by learning underlying (\ie task-independent) regularities.
\citet{thrun2012learning} generalize the classic definition of \emph{learning} given in~\citet{mitchell1997machine} to this setting and call
it \emph{learning to learn}.
Co-training on related tasks has become increasingly common over time and there have been many reports of improved generalization
as a result; \citet{collobert2008unified} is a notable example while~\citet{zhang2017survey} provides a recent survey.
Our present work can be seen as taking the approach~\citet{baxter1995learning} suggests to an extreme,
by pooling over tasks spanning a computationally-universal family of decision problems.

The phrase \emph{zero-shot learning} has been used in different contexts in the literature.
For example, the recent survey \citet{wang2019survey} defines it narrowly in terms of the multi-class classification problem
where some classes lack labels during training. We use it more broadly to mean success on new tasks not seen during training.
Besides~\cite{brown2020language},
the closest to our zero-shot results is perhaps~\citet{selsam2018learning},
which showed that training on a single synthetic distribution of satisfiability problems
permits zero-shot transfer to a suite of SAT problems encoding many diverse domains.
Central to the approach of \citet{selsam2018learning} was to use propositional
logic as the shared representation for all the tasks,
and as a result the applicability of their approach is severely limited by the inherent limitations of propositional logic.
Our present work uses a higher-order, computationally universal language for the shared representation and thus has no
such limitations.

Our \emph{universal policy} differs significantly from the AIXI theory of \emph{universal artificial intelligence}~\cite{hutter2000theory}.
AIXI is a learning agent that operates in an environment that is unknown and not software-defined;
it is universal in the sense that it maximizes expected future reward with respect to a \emph{universal}
prior probability distribution over environments, \ie one that assigns nonzero probability to all computable
hypotheses~\cite{solomonoff1964formal,solomonoff1964formalb}.
In contrast, our oracle is \emph{given} the source code of the environment,
and is universal in the sense that it can provide heuristic guidance for
whatever computable environment that it may be given.

Here we survey additional features that a more mature oracle-guided decision language may support.
We already mentioned \emph{dependency-directed backtracking} as developed by~\cite{zabih1987non}.
A second feature is \emph{partial evaluation}. The idea is that after some choices have been made,
it may be possible to simplify the continuation dramatically before making the next choice.
While dependency-directed backtracking can be seen as a generalization of conflict-driven back-jumping in SAT solvers,
partial evaluation can be seen as a generalization of unit propagation.
A third feature is the ability to decode entire sequences in a single query.
Currently Dodona may only query the oracle to decide among a finite set of possibilities;
to produce a list, it must query the oracle many times in sequence,
each time building the (only slightly evolving) continuation graph.
A fourth, as mentioned in \S\ref{subsec:dodona:tour}, is to support the full MDP family.
We could support arbitrary real-valued rewards at arbitrary times in Dodona by introducing a new effect \lstinline{(reward <float>)},
though since this effect does
not behave functionally, we would also need to introduce blocks into the language so that statements can be sequenced,
\ie using the \lstinline{(begin ...)} form from Scheme. Stochastic transitions can be added by simply adding
random primitives to the language. However, stochasticity would require additional meta-evaluators to handle
effectively, as best-first search would no longer be as sensible an operation.

We have presented a simple, stand-alone oracle-guided decision language, but
this paradigm can be realized in various ways within existing languages as well.
Ignoring for a moment the challenge of producing embeddable representations of the continuations,
nondeterminism can be simulated in a traditional language by writing programs
that \emph{explicitly} return choicepoints
where the continuations are simply functions in the language.
The catch is that this encoding requires an inordinate number of functions and
is syntactically cumbersome in traditional languages. However, nondeterminism constitutes
a monad, and this proliferation of functions is merely a special case of the proliferation of
functions when using any monad, which the \lstinline{do} notation pioneered by Haskell
is designed to hide~\cite{jones1995system,peterson1996report}. Embedding the continuation
poses a different challenge though, since if the continuation is represented as a function in the
base language, some significant meta-programming capabilities will be required to inspect it.
In \S\ref{sec:embed} we also made use of runtime type information to improve the
embedding, \eg by making the embedding for sets invariant to permutations of their elements.

The elephant in the room for oracle-guided decision programming is \emph{where will the training data come from}?
Language models have a clear advantage here: since the data is not required to have any particular structure or meaning,
almost any data will do.
The benefits of our approach that we discussed in \S\ref{sec:intro} come at the cost of requiring that we be more selective
about the data we train on. Ultimately we envision an OGDP framework embedded in a general-purpose programming language with
large-scale formal mathematics, software verification, and program synthesis efforts all sharing an oracle.
The oracle could also be seeded by training on synthetic data of all sorts.
It would even be possible to co-train it as a language model as well, by postulating an opaque \lstinline{next-token}
function that takes some data representing provenance (\eg the website the text was scraped from) along with
the usual sequence of tokens indicating the context.

Lastly, although the present work has focused on the universality of the oracle,
the broader paradigm of oracle-guided decision programming may also be
useful when used with domain-specific oracles.
The primitive \lstinline{choice} can be extended with a second argument that represents
some data that is to serve as input to a neural network of the user's choice.
Such a decision language could still provide value in the form of compositional ways
of building MDPs and generic (oracle-parameterized) search procedures.
Indeed, when using a domain-specific oracle, the MDP need not even be software-defined in the usual
sense; the program could be specialized for each individual input, \eg by including an empirically-observed
label for that input.

\section*{Acknowledgements}

We thank Todd Mytkowicz, Madan Musuvathi, and Percy Liang for helpful discussions.

\bibliography{universal}
\bibliographystyle{icml2020}

\end{document}

%% file: lstdodona.tex
\definecolor{scheme-choose}{rgb}{1, 0, 0}
\definecolor{scheme-primitive}{rgb}{0, 0, 1}
\definecolor{scheme-derived}{rgb}{0, 0, 1}
\definecolor{scheme-comment}{rgb}{0, 0.5, 0}
\definecolor{scheme-bracket}{rgb}{0.6, 0.6, 0.6}

\lstdefinelanguage{dodona}{
    basicstyle=\small\ttfamily,
    sensitive=true,
    alsoletter=-,
    morestring = [b]",
    stringstyle=\color{red},
    morecomment=[l]{;},
    commentstyle=\color{scheme-comment},
    literate=
            *{(}{{\textcolor{scheme-bracket}{(}}}{1}
            {)}{{\textcolor{scheme-bracket}{)}}}{1},
    classoffset=0,
    morekeywords={
      choose,
      fail,
      reward
      },
    keywordstyle=\color{scheme-choose},
    classoffset=1,
    morekeywords={
      eval,
        lambda,
        if,
        quote,
        set!},
    keywordstyle=\color{scheme-primitive},
    classoffset=2,
    morekeywords={
        cond,
        else,
        case,
        and,
        or,
        when,
        unless,
        cond-expand,
        let,
        let*,
        letrec,
        letrec*
        let-values,
        let*-values,
        begin,
        do,
        delay,
        delay-force,
        force,
        promise?,
        make-promise,
        make-parameter,
        parameterize,
        guard,
        quasiquote,
        case-lambda,
        let-syntax,
        letrec-syntax,
        syntax-rules,
        define,
        map},
    keywordstyle=\color{scheme-derived},
    classoffset=0
}